\documentclass{article}
\usepackage{ijcai18}

\usepackage{times}
\usepackage{xcolor}
\usepackage{soul}
\usepackage[utf8]{inputenc}
\usepackage[small]{caption}

\usepackage{tabularx}
\usepackage{xparse}
\usepackage{xspace}
\usepackage{relsize}
\usepackage{graphicx}
\usepackage{multirow}
\usepackage[normalem]{ulem}
\usepackage{amsmath}
\usepackage{xr}
\usepackage{url}
\usepackage{booktabs}
\usepackage{xcolor}
\usepackage{todonotes}
\usepackage{multirow}
\usepackage{colortbl}
\usepackage{setspace}

\usepackage{subfigure}
\def\bb{\mathbf{b}}
\def\bc{\mathbf{c}}

\def\bh{\mathbf{h}}

\def\by{\mathbf{y}}

\def\bq{\mathbf{q}}

\def\bx{\mathbf{x}}

\def\by{\mathbf{y}}

\DeclareMathOperator{\softmax}{\mathrm{softmax}}
\allowdisplaybreaks

\newenvironment{itemize*}%
  {\begin{itemize}%
    \setlength{\itemsep}{0pt}%
    \setlength{\parskip}{0pt}}%
  {\end{itemize}}
  \newenvironment{enumerate*}%
  {\begin{enumerate}%
    \setlength{\itemsep}{0pt}%
    \setlength{\parskip}{0pt}}%
  {\end{enumerate}}

\begin{document}
% The file aaai.sty is the style file for AAAI Press
% proceedings, working notes, and technical reports.
%
\title{Same Representation, Different Attentions:\\ A New Scheme of Information Sharing for Multi-task Learning}
\title{Same Representation, Different Attentions:\\Shareable Sentence  Representation Learning from Multiple Tasks}
\author{Renjie Zheng, Junkun Chen, Xipeng Qiu\thanks{Corresponding Author, xpqiu@fudan.edu.cn}\\
Shanghai Key Laboratory of Intelligent Information Processing, Fudan University\\
School of Computer Science, Fudan University\\
825 Zhangheng Road, Shanghai, China\\
}

\maketitle

\begin{abstract}
%\begin{quote}
Distributed representation plays an important role in deep learning based natural language processing.   However, the representation of a sentence often varies in different tasks, which is usually learned from scratch and suffers from the limited amounts of training data. In this paper, we claim that a good sentence representation should be invariant and can benefit the various subsequent tasks. To achieve this purpose, we propose a new scheme of information sharing for multi-task learning. More specifically, all tasks share the same sentence representation and each task can select the task-specific information from the shared sentence representation with attention mechanism. The query vector of each task's attention could be either static parameters or generated dynamically.
We conduct extensive experiments on 16 different text classification tasks, which demonstrate the benefits of our architecture.
%Moreover, our method can introduce a different type of task as auxiliary task and the performance can be further boosted.
%\end{quote}
\end{abstract}

\section{Introduction}\label{sec:intro}
The distributed representation plays an important role in deep learning based natural language processing (NLP) \cite{bengio2003neural,collobert2011natural,sutskever2014sequence}. On word level, many successful methods have been proposed to learn a good representation for single word, which is also called word embedding, such as skip-gram \cite{mikolov2013distributed}, GloVe \cite{pennington2014glove}, etc. There are also pre-trained word embeddings, which can easily used in downstream tasks. However, on sentence level, there is still no generic sentence representation which is suitable for various NLP tasks.
%A good representation of the variable-length text should fully capture the semantics of natural language. %These models can be learned by supervised or unsupervised way.

Currently, most of sentence encoding models are trained specifically for a certain task in a supervised way, which results to different representations for the same sentence in different tasks.
Taking the following sentence as an example for domain classification task and sentiment classification task,
\begin{center}
\begin{tabular}{l}
  \textit{The \textbf{infantile cart} is \textbf{easy to use}, }
\end{tabular}
\end{center}
general text classification models always learn two representations separately.
For domain classification, the model can learn a better representation of ``\textit{infantile cart}'' while for sentiment classification, the model is able to learn a better representation of ``\textit{easy to use}''.
%we learn its representation separately in two tasks: domain classification and sentiment classification.
%Obviously, the sentence encoding model of each task just need extract the task-specific information. Thus, the model of domain classification can only learn the representation of ``\textit{infantile cart}'' and ignores ``\textit{easy to use}''. Otherwise, the model of sentiment classification can only learn the representation of ``\textit{easy to use}'' and ignores ``\textit{infantile cart}''.

%Although the task-specific sentence representation is usually better than generic task-independent ones, they are hard to be transferred to other tasks.
However, to train a good task-specific sentence representation from scratch, we always need to prepare a large dataset which is always unavailable or costly.
To alleviate this problem, one approach is pre-training the model on large unlabeled corpora by unsupervised learning tasks, such as language modeling \cite{bengio2003neural}. This unsupervised pre-training may be helpful to improve the final performance, but the improvement is not guaranteed since it does not directly optimize the desired task.

\begin{figure}[t]
\begin{center}
\subfigure[]{\label{fig:illustration1}
\includegraphics[height=3cm]{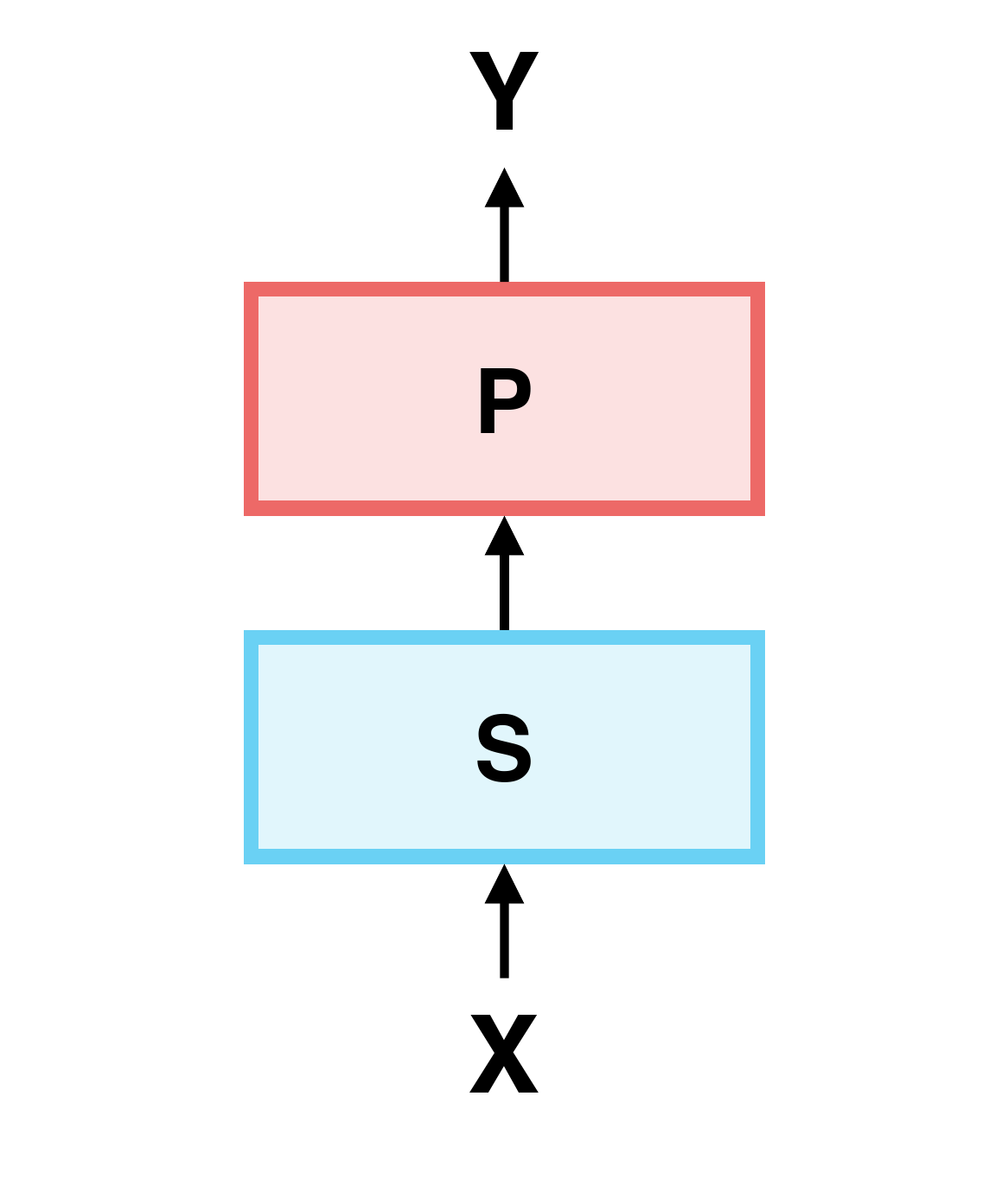}
}
\subfigure[]{\label{fig:illustration2}
\includegraphics[height=3cm]{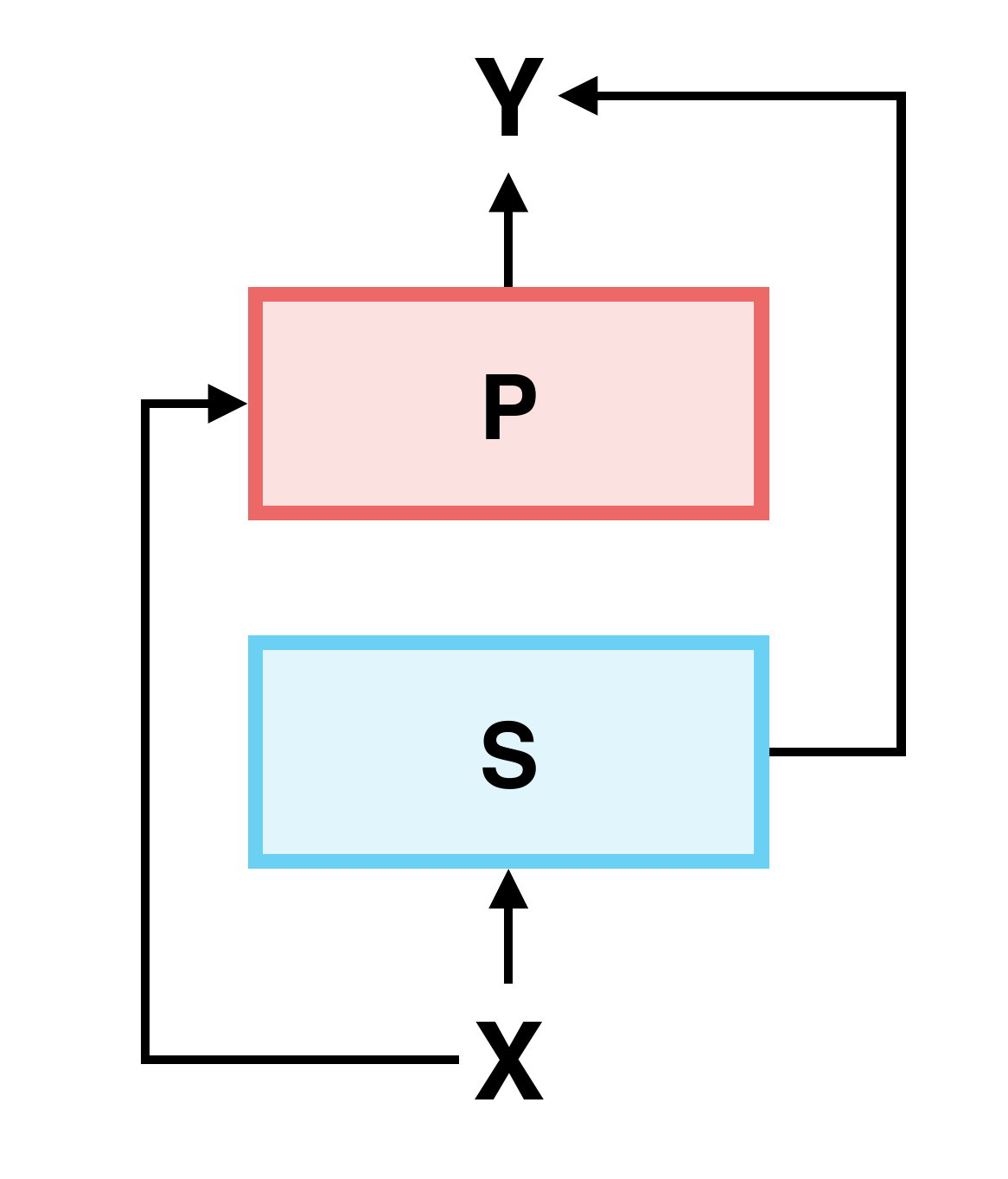}
}
\subfigure[]{\label{fig:illustration3}
\includegraphics[height=3cm]{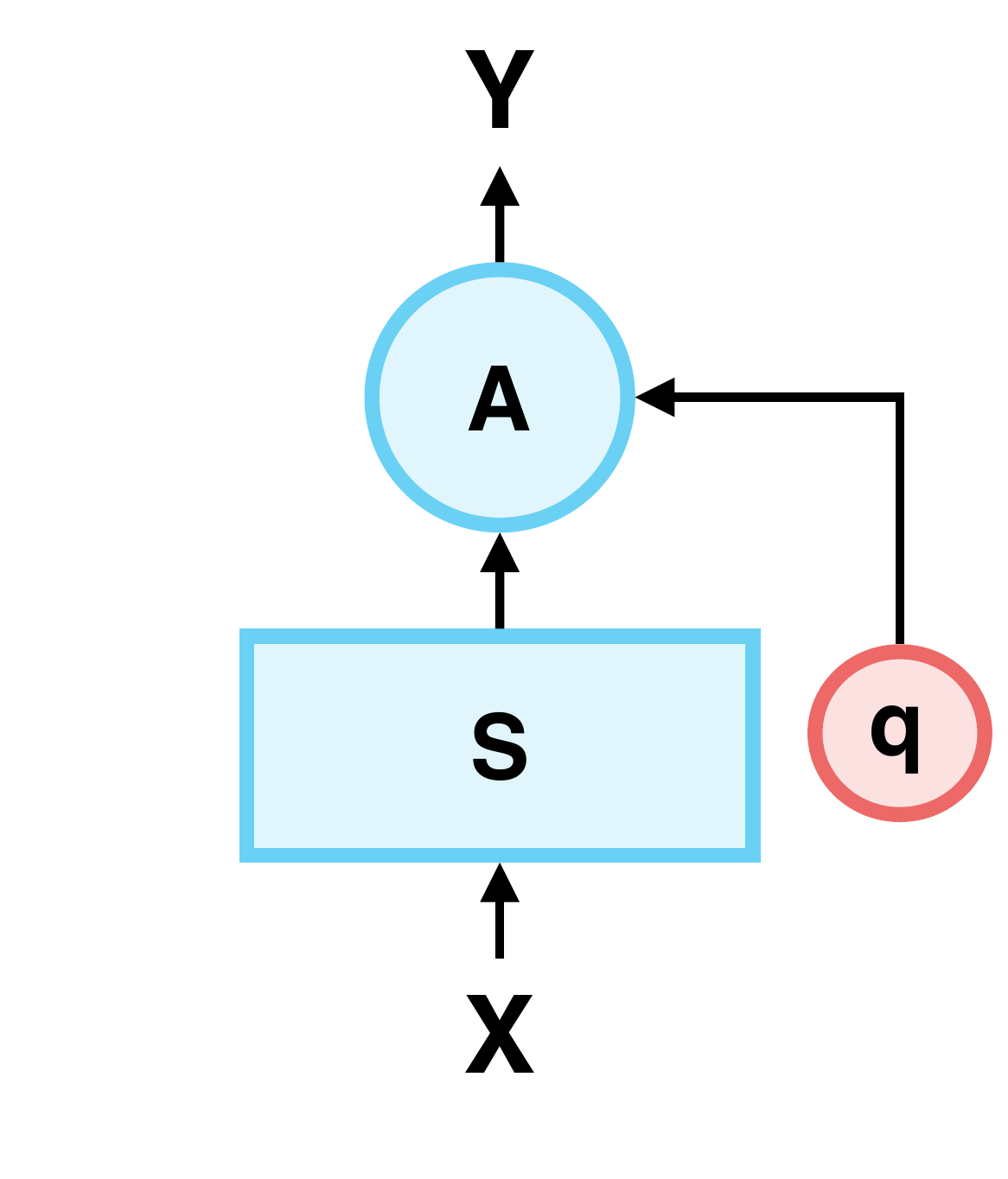}
}
\end{center}
%\vspace{-3ex}
\caption{Three schemes of information sharing in multi-task leaning. (a) stacked shared-private scheme, (b) parallel shared-private scheme, (c) our proposed attentive sharing scheme.}
\vspace{-3ex}
\label{fig:illustration}
\end{figure}

Another approach is multi-task learning  \cite{caruana1997multitask}, which is an effective approach to improve the performance of a single task with the help of other related tasks. However, most existing models on multi-task learning attempt to divide the representation of a sentence into private and shared spaces. The shared representation is used in all tasks, and the private one is different for each task.
The two typical information sharing schemes are stacked shared-private scheme and parallel shared-private scheme (as shown in Figure \ref{fig:illustration1} and \ref{fig:illustration2} respectively).
%However, a major challenge in these models is how to design the mechanism of information sharing among the different tasks to avoid the problem of \textit{negative transfer}.
%Negative transfer is that improper sharing information between two tasks can worsen the performance of both tasks.
However, we cannot guarantee that a good sentence encoding model is learned by the shared layer.

To learn a better shareable sentence representation, we propose a new information-sharing scheme for multi-task learning in this paper. In our proposed scheme, the representation of every sentence is fully shared among all different tasks.  To extract the task-specific feature, we utilize the attention mechanism and introduce a task-dependent query vector to select the task-specific information from the shared sentence representation. The query vector of each task can be regarded as learnable parameters (static) or be generated dynamically.
If we take the former example, in our proposed model these two classification tasks share the same representation which includes both domain information and sentiment information. On top of this shared representation, a task-specific query vector will be used to focus ``\textit{infantile cart}'' for domain classification and ``\textit{easy to use}'' for sentiment classification.
%The task-specific information is just selected from the shared representation, rather than the secondary process of the shared representation. Thus, we can substantially learn a good representation with the help of the data of different tasks.

The contributions of this papers can be summarized as follows.
\begin{itemize*}
\item We propose a new information sharing scheme for multi-task learning. As a side effect, the model can be easily visualized and shows what specific parts of the sentence are focused in different tasks.
\item In our proposed scheme, we can learn a shareable generic sentence representation, which can be easily transferred to other tasks. The shareable sentence representation can also be improved by the auxiliary tasks, such as POS Tagging and Chunking.
\item We conduct extensive experiments on 16 sentiment classification tasks. Experiments show that our proposed model is space efficient and converges quickly.
\end{itemize*}

\section{Sentence Encoding in Multi-task Learning}\label{sec:method}

\subsection{Neural Sentence Encoding Model}
The primary role of sentence encoding models is to represent the variable-length sentence or paragraphs as fixed-length dense vector (distributed representation).
Currently, the effective neural sentence encoding models include neural Bag-of-words (NBOW), recurrent neural networks (RNN) \cite{sutskever2014sequence,chung2014empirical}, convolutional neural networks (CNN) \cite{collobert2011natural,kalchbrenner2014convolutional,kim2014convolutional}, and syntactic-based compositional model \cite{socher2013recursive,tai2015improved,zhu2015long}.

Given a text sequence $x = \{x_1, x_2, \cdots, x_T\}$, we first use a lookup layer to get the vector representation (word embedding) $\bx_i$ of each word $x_i$. Then we can use CNN or RNN to calculate the hidden state $\bh_i$ of each position $i$.
The final representation of a sentence could be either the final hidden state of the RNN or the max (or average) pooling from all hidden states of RNN (or CNN).

%Here we adopt recurrent neural network with long short-term memory (LSTM) due to their superior performance in various NLP tasks.

% \paragraph{Long Short-term Memory} Long short-term memory network (LSTM) \cite{hochreiter1997long} is a type of recurrent neural network (RNN) \cite{elman1990finding}, and specifically addresses the issue of learning long-term dependencies. We define the LSTM units at each time step $t$ to be a collection of vectors in $R^d$ : an input gate $i_t$ , a forget gate $f_t$ , an output gate $o_t$ , a memory cell $c_t$ and a hidden state $h_t$ . d is the number of the LSTM units. The elements of the gating vectors $i_t$ , $f_t$ and $o_t$ are in [0, 1].

% The LSTM is precisely specified as follows.
% \begin{align}
% \begin{split}\label{eq:1}
% 	\begin{bmatrix}
% 		\tilde{\bc_t} \\ \bo_t \\ \bi_t \\ \mathbf{f}_{t}
% 	\end{bmatrix}
% 	&=
% 	\begin{bmatrix}
% 		tanh \\ \sigma \\ \sigma \\ \sigma
% 	\end{bmatrix}
% 	 (W_p \begin{bmatrix} \bx_t \\ \bh_{t-1} \end{bmatrix} + \bb_p)
% \end{split}\\
% \begin{split}\label{eq:2}
%  \bc_t ={}& \tilde{\bc}_t \odot \bi_t + \bc_{t-1} \odot \mathbf{f}_{t},
% \end{split}\\
%  \bh_t ={}& \bo_t \odot tanh (\bc_t),
% \end{align}
% where $\bx_t \in R^e$ is the input at the current time step; $W_p \in R^{4d \times (d+e)}$ and $\bb_p \in R^{4d}$ are parameters of affine transformation; $\sigma$ denotes the logistic sigmoid function and $\odot$ denotes element-wise multiplication.

We use bidirectional LSTM (BiLSTM) to gain some dependency between adjacent words.
The update rule of each LSTM unit can be written as follows:
 \begin{align}
 \overrightarrow{\bh_t} &= \rm{LSTM}( \overrightarrow{\bh}_{t-1}, \bx_t, \theta_p),\\
 \overleftarrow{\bh_t} &= \rm{LSTM}( \overleftarrow{\bh}_{t+1}, \bx_t, , \theta_p),\\
 {\bh} &= \frac{1}{T}\sum_{t=1}^{T} \overrightarrow{\bh_t}\oplus\overleftarrow{\bh_t},
 \end{align}
where $\theta_p$ represents all the parameters of BiLSTM. The representation of the whole sequence is the average of the hidden states of all the positions, where $\oplus$ denotes the concatenation operation.

\if 0
The representation of the whole sequence is the average of the hidden states of all the positions, which has a fully connected layer followed by a softmax non-linear layer that predicts the probability distribution over classes.

%\begin{align}
%{\bh} = \frac{1}{T}\sum_{t=1}^{T} \overrightarrow{\bh_t}\oplus\overleftarrow{\bh_t},
%\end{align}

where $\oplus$ denotes the concatenation operation.
\fi

\subsection{Shared-Private Scheme in Multi-task Learning}

Multi-task Learning \cite{caruana1997multitask} utilizes the correlation between related tasks to improve classification by learning tasks in parallel, which has been widely used in various natural language processing tasks, such as text classification \cite{liu2016recurrent}, semantic role labeling \cite{collobert2008unified}, machine translation \cite{firat2016multi}, and so on.

%The goal of multi-task learning is to utilizes the correlation among these related tasks to improve classification by learning tasks in parallel.
To facilitate this, we give some explanation for notations used in this paper.
Formally, we refer to $\mathcal{D}_k$ as a dataset with $N_k$ samples for task $k$. Specifically,
\begin{equation}
D_k = \{(x_i^{(k)},y_i^{(k)})\}_{i=1}^{N_k}
\end{equation}
where $x_i^{(k)}$ and $y_i^{(k)}$ denote a sentence and corresponding label for task $k$.

\paragraph{Shared-Private Scheme} A common information sharing scheme is to divide the feature spaces into two parts: one is used to store task-specific features, the other is used to capture task-invariant features. As shown in Figure \ref{fig:illustration1} and \ref{fig:illustration2}, there are two schemes: stacked shared-private (SSP) scheme and parallel shared-private (PSP) scheme.

In stacked scheme, the output of the shared LSTM layer is fed into the private LSTM layer, whose output is the final task-specific sentence representation.
In parallel scheme, the final task-specific sentence representation is the concatenation of outputs from the shared LSTM layer and the private LSTM layer.

%\begin{figure}[t]
%\centering
%\includegraphics[height=3.5cm]{img/sp-mtl.png}
%\caption{Shared-private scheme in multi-task learning.}
%\label{fig:sp-mtl}
%\end{figure}

\paragraph{Task-Specific Output Layer}

For a sentence $x^{(k)}$ and its label $y^{(k)}$ in task $k$, its final representation is ultimately fed into the corresponding task-specific softmax layer for classification or other tasks.
\begin{align}
{\hat{\by}^{(k)}} = \softmax(W^{(k)} \bh^{(k)} + \bb^{(k)})\label{eq:softmax}
\end{align}
where ${\hat{\by}^{(k)}}$ is prediction probabilities; $\bh^{(k)}$ is the final task-specific representation; $W^{(k)}$ and $\bb^{(k)}$ are task-specific weight matrix and bias vector respectively.

The total loss $L_{task}$ can be computed as:
\begin{align}
\mathcal{L}_{All} &= \sum_{k=1}^{K}{\alpha}_k  \mathcal{L}_{Task}({\hat{y}}^{(k)}, y^{(k)})
\end{align}
where $\alpha_k$ (usually set to 1) is the weights for each task $k$ respectively; $\mathcal{L}_{Task}({\hat{y}}, y)$ is the cross-entropy of the predicted and true distributions.
%The parameters of the network are trained to minimize the total loss $\mathcal{L}_{All}$ on all the tasks.

\section{A New Information-Sharing Scheme for Multi-task Learning}

The key factor of multi-task learning is the information sharing scheme in latent representation space. Different from the traditional shared-private scheme, we introduce a new scheme for multi-task learning on NLP tasks, in which the sentence representation is shared among all the tasks, the task-specific information is selected by attention mechanism.

%\subsection{Task-Attentive Sentence Encoding Model}
%In standard neural network which consist of a series of non-linear transformation layers, it is hard for it to process local information. For example in sentiment analysis, the most differential part of the classification of given text are those sentiment words. This problem is very important for long text.
In a certain task, not all information of a sentence is useful for the task, therefore we just need to select the key information from the sentence. Attention mechanism \cite{bahdanau2014neural,mnih2014recurrent} is an effective method to select related information from a set of candidates. The attention mechanism can effectively solve the capacity problem of sequence models, thereby is widely used in many NLP tasks, such as machine translation \cite{luong2015effective}, textual entailment \cite{zhao2017textual} and summarization \cite{rush2015neural}.

\subsection{Static Task-Attentive Sentence Encoding}

\begin{figure}[t]
\begin{center}
\subfigure{
\includegraphics[width=0.4\textwidth]{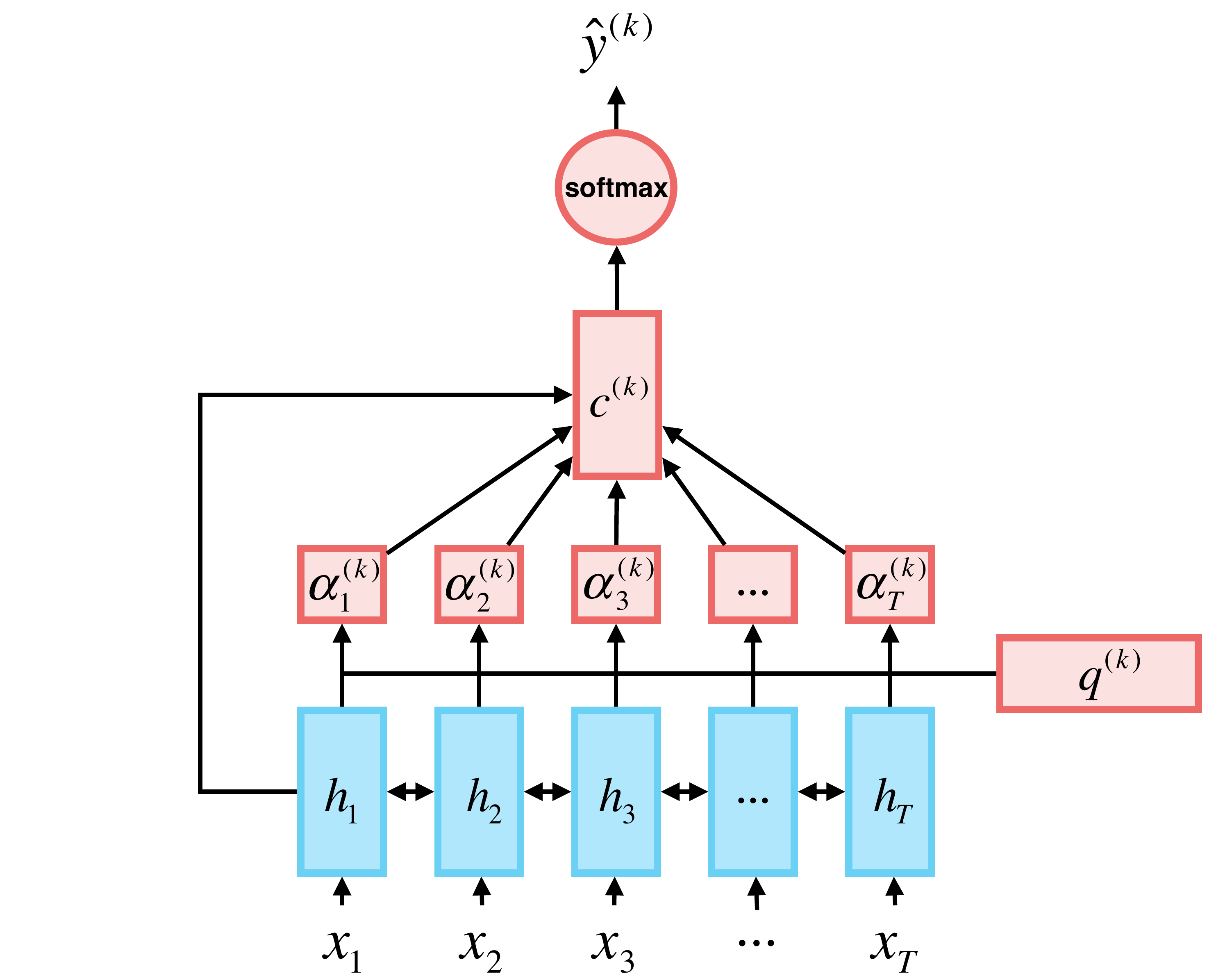}
}
\vspace{-3ex}
\end{center}
\caption{Static Task-Attentive Sentence Encoding}\label{fig:static}
\vspace{-2ex}
\end{figure}

We first introduce the static task-attentive sentence encoding model, in which the task query vector is a static learnable parameter. As shown in Figure \ref{fig:static}, our model consists of one shared BiLSTM layer and an attention layer. Formally, for a sentence in task $k$, we first use BiLSTM to calculate the shared representation $[\bh_1,\cdots,\bh_T]$. Then we use attention mechanism to select the task-specific information from a generic task-independent sentence representation. Following \cite{luong2015effective}, we use the dot-product attention to compute the attention distribution. We introduce a task-specific query vector $\bq^{(k)}$ to calculate the attention distribution $\alpha^{(k)}$ over all positions.
\begin{align}
%\alpha^{(k)}_t &= \frac{\exp({\bq^{(k)}}^T \bh_t)}{\sum_{t'=1}^{T} \exp({\bq^{(k)}}^T \bh_{t'})},
\alpha^{(k)}_t &= \softmax({\bq^{(k)}}^T \bh_t),
\end{align}
where the task-specific query vector $\bq^{(k)}$ is a learned parameter. The final task-specific representation $\bc^{(k)}$ is summarized by
\begin{align}
\bc^{(k)} &= \sum_{t = 1}^T \alpha^{(k)}_t \bh_t.
\end{align}

At last, a task-specific fully connected layer followed by a softmax non-linear layer processes the task-specific context $\bc^{(k)}$ and predicts the probability distribution over classes.

\begin{figure}[t]
\centering
	\includegraphics[width=0.22\textwidth]{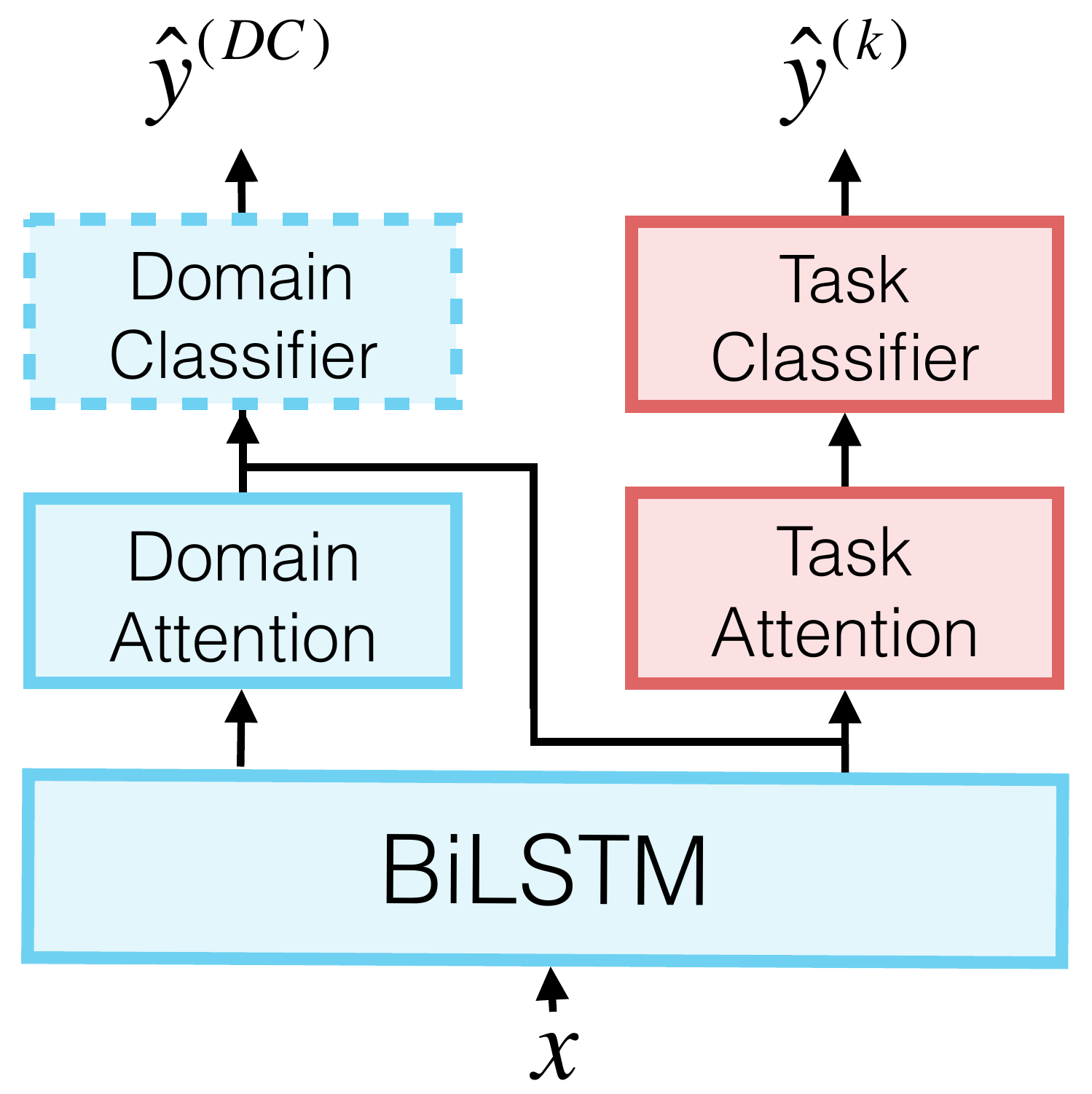}
\vspace{-1ex}
\caption{
Dynamic Task-Attentive Sentence Encoding
}
\label{fig:model}
\vspace{-2ex}
\end{figure}

\subsection{Dynamic Task-Attentive Sentence Encoding}  Different from the static task-attentive sentence encoding model, the query vectors of the dynamic task-attentive sentence encoding model are generated dynamically. When each task belongs to a different domain, we can introduce an auxiliary domain classifier to predict the domain (or task) of the specific sentence. Thus, the domain information is also included in the shared sentence representation, which can be used to generate the task-specific query vector of attention.

The original tasks and the auxiliary task of domain classification (DC) are joint learned in our multi-task learning framework.

The query vector  $\bq^{(DC)}$ of DC task is static and needs be learned in training phrase. The domain information is also selected with attention mechanism.
\begin{align}
\alpha^{(DC)}_t &= \softmax({\bq^{(DC)}}^T \bh_t),\\
\bc^{(DC)} &= \sum_{t = 1}^T \alpha^{(DC)}_t \bh_t.\\
\hat{y}^{(DC)} &= \rm{softmax}(W^{(DC)} \bc^{(DC)} + \bb^{(DC)}),
\end{align}
where $\alpha^{(DC)}$ is attention distribution of auxiliary DC task, and $\bc^{(DC)}$ is the attentive information for DC task, which is fed into the final classifier to predict its domain $\hat{y}^{(DC)}$.

Since $\bc^{(DC)}$ contains the domain information, we can use it to generate a more flexible query vector
\begin{align}
\bq^{(k)} = U \bc^{(DC)} + \bb^{(k)},
\end{align}
where $U$ is a shared learnable weight matrix and $\bb^{(k)}$ is a task-specific bias vector. When we set $U=0$, the dynamic query is equivalent to the static one.

%Those two classifiers are trained jointly using the same corpus but with different classification label: domain label for the domain classification corpus and sentiment label for the text classification corpus. The domain classifier is a classic text classification model with attention network as we introduce in the previous section.

\section{Experiment}\label{sec:experiment}
In this section, we investigate the empirical performances of our proposed architectures on three experiments.

\subsection{Exp I: Sentiment Classification}
We first conduct a multi-task experiment on sentiment classification.

\paragraph{Dataset} We use 16 different datasets %\footnote{\url{http://pfliu.com/paper/adv-mtl.html}}
from several popular review corpora used in \cite{liu2017adversarial}. These datasets consist of 14 product review datasets and two movie review datasets.
%The product review datasets \cite{blitzer2007biographies} are collected from Amazon product reviews from different domains, such as Books, DVDs, Electronics, etc. The movie review datasets contains the IMDB dataset \cite{maas2011learning} and the MR dataset \cite{pang2005seeing}.

All the datasets in each task are partitioned randomly into training set, development set and testing set with the proportion of 70\%, 10\% and 20\% respectively. The detailed statistics about all the datasets are listed in Table \ref{tab:dataset}.

\begin{table}[t]\setlength{\tabcolsep}{5pt}\small
\centering
\begin{tabular*}{0.5\textwidth}{l @{\extracolsep{\fill}} rrrrrr} \hline
\textbf{Dataset} & Train & Dev. & Test & Avg. L & Vocab. \\ \hline
Books & 1400 & 200 & 400 & 159 & 62K \\
Elec. & 1398 & 200 & 400 & 101 & 30K \\
DVD & 1400 & 200 & 400 & 173 & 69K \\
Kitchen & 1400 & 200 & 400 & 89 & 28K \\
Apparel & 1400 & 200 & 400 & 57 & 21K \\
Camera & 1397 & 200 & 400 & 130 & 26K \\
Health & 1400 & 200 & 400 & 81 & 26K \\
Music & 1400 & 200 & 400 & 136 & 60K \\
Toys & 1400 & 200 & 400 & 90 & 28K \\
Video & 1400 & 200 & 400 & 156 & 57K \\
Baby & 1300 & 200 & 400 & 104 & 26K \\
Mag. & 1370 & 200 & 400 & 117 & 30K \\
Soft. & 1315 & 200 & 400 & 129 & 26K \\
Sports & 1400 & 200 & 400 & 94 & 30K \\
IMDB & 1400 & 200 & 400 & 269 & 44K \\
MR & 1400 & 200 & 400 & 21 & 12K \\ \hline
\end{tabular*}
\caption{Statistics of the 16 datasets. The columns 2-5 denote the number of samples in training, development and test sets. The last two columns represent the average length and vocabulary size of the corresponding dataset.}
\vspace{-3ex}
\label{tab:dataset}
\end{table}

\paragraph{Competitor Methods}
We compare our proposed two information sharing schemes, static attentive sentence encoding (SA-MTL) and dynamic attentive sentence encoding (DA-MTL), with the following multi-task learning frameworks.
\begin{itemize*}
    \item \textbf{FS-MTL}: This model is a combination of a fully shared BiLSTM and a classifier.
    \item \textbf{SSP-MTL}: This is the stacked shared-private model as shown in Figure \ref{fig:illustration1} whose output of the shared BiLSTM layer is fed into the private BiLSTM layer.
    \item \textbf{PSP-MTL}: The is the parallel shared-private model as shown in Figure \ref{fig:illustration2}. The final sentence representation is the concatenation of both private and shared BiLSTM.
    \item \textbf{ASP-MTL}: This model is proposed by \cite{liu2017adversarial} based on PSP-MTL with uni-directional LSTM. The model uses adversarial training to separate task-invariant and task-specific features from different tasks.
\end{itemize*}
\vspace{-2ex}

\paragraph{Hyperparameters} We initialize word embeddings with the 200d GloVe vectors (840B token version, \cite{pennington2014glove}). The other parameters are initialized by randomly sampling from uniform distribution in [-0.1, 0.1]. The mini-batch size is set to 32. For each task, we take hyperparameters which achieve the best performance on the development set via a small grid search. We use ADAM optimizer \cite{kingma2014adam} with the learning rate of $0.001$. The BiLSTM models have 200 dimensions in each direction, and dropout with probability of $0.5$. During the training step of multi-task models, we select different tasks randomly. After the training step, we fix the parameters of the shared BiLSTM and fine tune every task.

\begin{table*}[t]\small
\centering\setlength{\tabcolsep}{5pt}
\begin{tabular}{l @{\extracolsep{\fill}} cccccccccc} \hline
\textbf{Task} & \multicolumn{3}{c}{Single Task} & \multicolumn{6}{c}{Multiple Tasks} \\
\cmidrule(lr){2-4} \cmidrule(lr){5-10}
            & BiLSTM&att-BiLSTM& Avg.&FS-MTL&SSP-MTL&PSP-MTL&ASP-MTL*&SA-MTL&DA-MTL \\ \hline
Books       & 81.0  & 82.0  & 81.5 & 84.0 & 85.5 & 85.5   & 87.0 & 86.8   & 88.5    \\
Electronics & 81.8  & 83.0  & 82.4 & 84.8 & 86.8 & 87.3   & 89.0 & 87.5   & 89.0    \\
DVD         & 83.3  & 83.0  & 83.1 & 85.0 & 85.3 & 84.5   & 87.4 & 87.3   & 88.0    \\
Kitchen     & 80.8  & 80.3  & 80.5 & 87.0 & 86.5 & 87.5   & 87.2 & 89.3   & 89.0    \\
Apparel     & 87.5  & 86.5  & 87.0 & 86.8 & 85.3 & 85.8   & 88.7 & 87.3   & 88.8    \\
Camera      & 87.0  & 89.5  & 88.3 & 89.0 & 90.5 & 90.3   & 91.3 & 90.3   & 91.8    \\
Health      & 87.0  & 84.3  & 83.0 & 88.5 & 88.3 & 87.5   & 88.1 & 88.3   & 90.3    \\
Music       & 81.8  & 82.0  & 81.8 & 81.0 & 84.5 & 83.0   & 82.6 & 84.0   & 85.0    \\
Toys        & 81.5  & 85.0  & 85.4 & 88.3 & 87.0 & 87.8   & 88.8 & 89.3   & 89.5    \\
Video       & 83.0  & 83.5  & 83.3 & 85.0 & 87.3 & 88.0   & 85.5 & 88.5   & 89.5    \\
Baby        & 86.3  & 86.0  & 86.1 & 89.0 & 88.3 & 90.0   & 89.8 & 88.8   & 90.5    \\
Magazine    & 92.0  & 92.0  & 92.0 & 92.0 & 92.3 & 92.8   & 92.4 & 92.0   & 92.0    \\
Software    & 84.5  & 83.0  & 83.8 & 86.3 & 88.5 & 90.3   & 87.3 & 89.3   & 90.8    \\
Sports      & 86.0  & 84.8  & 85.4 & 88.3 & 88.8 & 86.8   & 86.7 & 89.8   & 89.8    \\
IMDB        & 82.5  & 83.5  & 83.0 & 82.3 & 84.0 & 84.5   & 85.8 & 87.5   & 89.8    \\
MR          & 74.8  & 76.0  & 75.4 & 71.3 & 70.8 & 69.0   & 77.3 & 73.0   & 75.5    \\ \hline \rowcolor[gray]{.9} AVG.
            & 83.7  & 84.0  & 83.9 &85.5(+1.6)& 86.2(2.3) & 86.2 (+2.3)  & 87.2(+3.3) & 87.6 (+3.7)   & 88.2 (+4.3)    \\ \hline
\# Param. & 644 K $\times$ 16  &  645 K $\times$ 16   & --   & 644 K & 16,074 K & 10,972 K & 5,490K & 668 K & 818 K \\ \hline
%\# Private Param. & -- &  --  & --   & 0 & 16,074 K & 10,972 K & xxx & 668 K & 818 K \\ \hline
\end{tabular}
\caption{Performances on 16 tasks. The column of ``Single Task'' includes bidirectional LSTM (BiLSTM), bidirectional LSTM with attention (att-BiLSTM) and the average accuracy of the two models. The column of ``Multiple Tasks'' shows several multi-task models. * is from
\protect\cite{liu2017adversarial}
.}
\vspace{-3ex}
\label{tab:result}
\end{table*}

\paragraph{Results}
Table \ref{tab:result} shows the performances of the different methods.  From the table, we can see that the performances of most tasks can be improved with the help of multi-task learning. FS-MTL shows the minimum performance gain from multi-task learning since it puts all private and shared information into a unified space. SSP-MTL and PSP-MTL achieve similar performance and are outperformed by ASP-MTL which can better separate the task-specific and task-invariant features by using adversarial training. Our proposed models (SA-MTL and DA-MTL) outperform ASP-MTL because we model a richer representation from these 16 tasks. Compared to SA-MTL, DA-MTL achieves a further improvement of $+0.6$  accuracy with the help of  the dynamic and flexible query vector. It is noteworthy that our models are also space efficient since the task-specific information is extracted by using only a query vector, instead of a BiLSTM layer in the shared-private models.

\begin{figure}[t]
\centering
\includegraphics[width=0.35\textwidth]{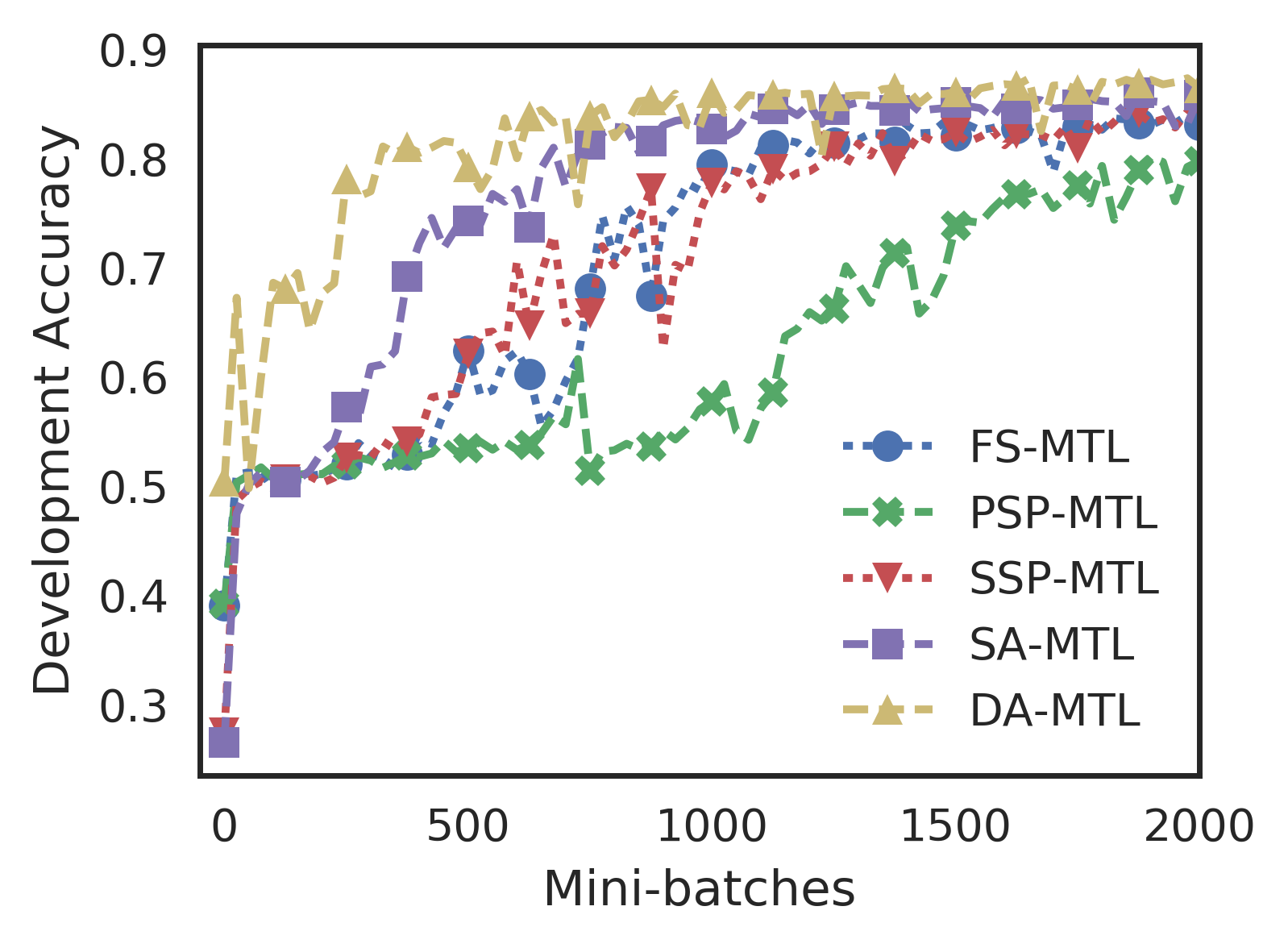}
\caption{Convergence on the development datasets.}
\label{fig:accracy}
\end{figure}

We also present the convergence properties of our models on the development datasets compared to other multi-task models in Figure \ref{fig:accracy}. We can see that PSP-MTL converges much more slowly than the rest four models because each task-specific classifier should consider the output of shared layer which is quite unstable during the beginning of training phrase. Moreover, benefit from the attention mechanism which is useful in feature extraction, SA-TML and DA-MTL are converged much more quickly than the rest of models.

\paragraph{Visualization}

\begin{figure}[t]
\begin{center}
\subfigure{
\includegraphics[height=5.5cm]{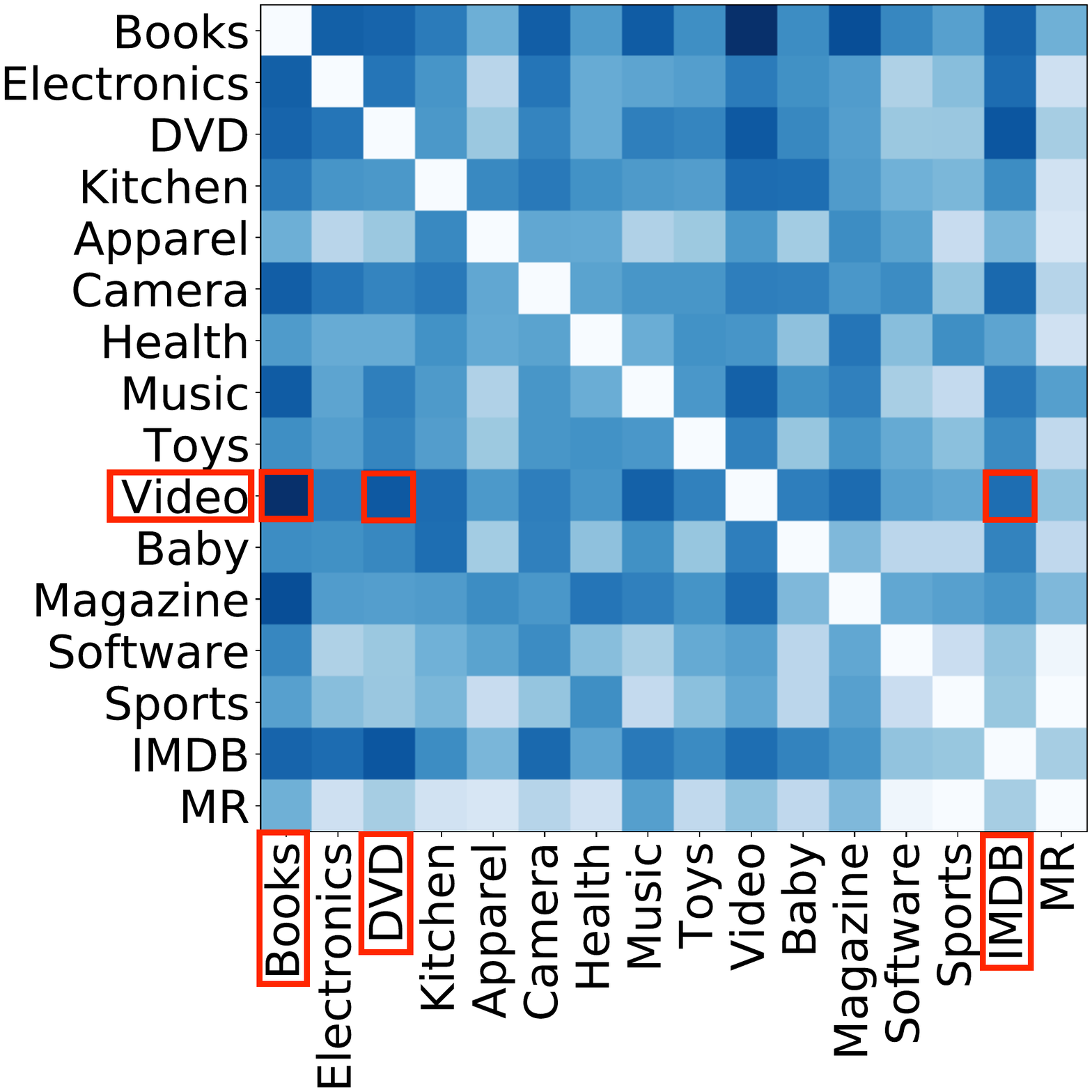}
}
\end{center}
\vspace{-3ex}
\caption{Similarity Matrix of Different Task's query vector $q_k$}
\vspace{-3ex}
\label{fig:q_matrix}
\end{figure}

Since all the tasks share the same sentence encoding layer, the query vector $\bq$ of each task determines which part of the sentence to attend. Thus, similar tasks should have the similar query vectors. Here we simply calculate the Frobenius norm of each pair of tasks' $\bq$ as the similarity.
Figure \ref{fig:q_matrix} shows the similarity matrix of different task's query vector $\bq$ in static attentive model. A darker cell means the higher similarity of the two task's $\bq$. Since the cells in the diagnose of the matrix denotes the similarity of one task, we leave them blank because they are meaningless. It's easy to find that $\bq$ of ``DVD'', ``Video'' and ``IMDB'' have very high similarity. It makes sense because they are all reviews related to movie. However, another movie review ``MR'' has very low similarity to these three task. It's probably that the text in ``MR'' is very short that makes it different from these tasks. The similarity of $\bq$ from ``Books'' and ``Video'' is also very high because these two datasets share a lot of similar sentiment expressions.

\begin{figure}[t]
\begin{tabular}{c}%\hspace{-2ex}
\begin{minipage}[t]{0.5\linewidth}
\begin{center}
\subfigure[\scriptsize{Attention of task ``Books'' in SA-MTL, Output: Negative}]{
\includegraphics[width=8cm]{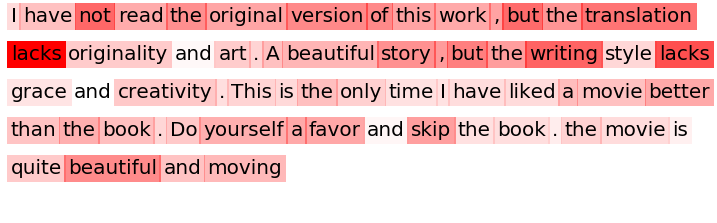}
}
\end{center}
\end{minipage}
\\
\begin{minipage}[t]{0.5\linewidth}
\begin{center}
\subfigure[\scriptsize{Attention of task ``DVD'' in SA-MTL, Output: Positive}]{
\includegraphics[width=8cm]{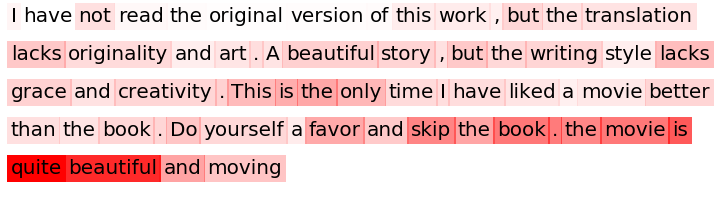}
}
\end{center}%\vspace{-4ex}
\end{minipage}
\\
\begin{minipage}[t]{0.5\linewidth}
\begin{center}
\subfigure[\scriptsize{Attention of auxiliary Task (Domain Classification) in DA-MTL, Ouptut: Books}]{
\includegraphics[width=8cm]{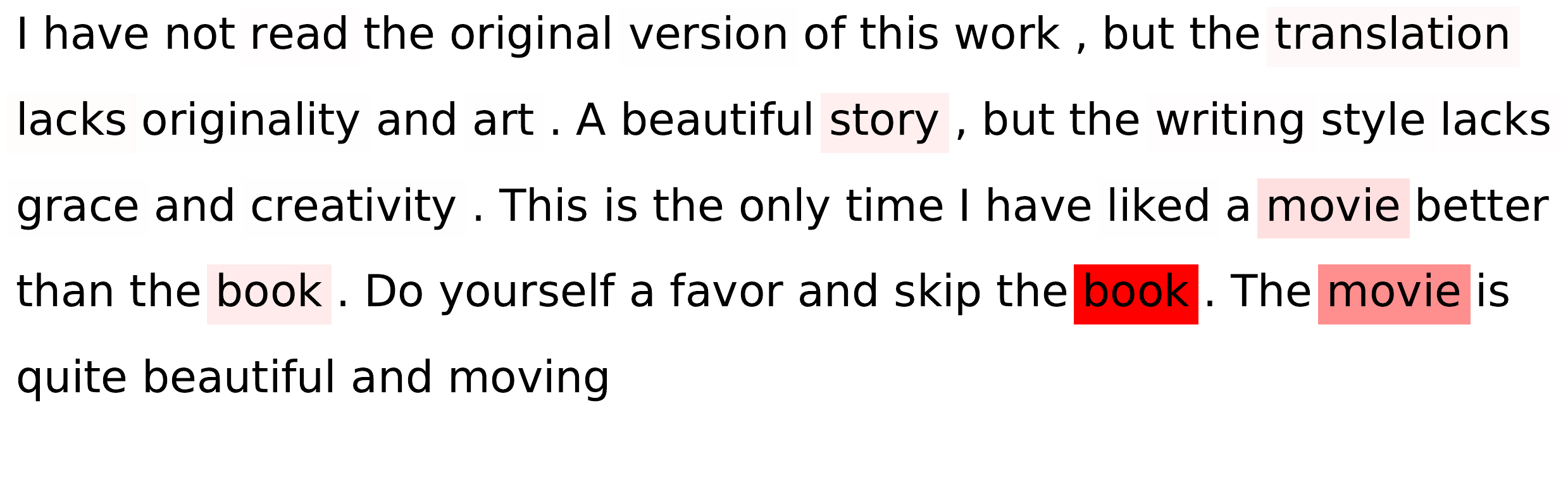}
}
\end{center}%\vspace{-4ex}
\end{minipage}
\\
\begin{minipage}[t]{0.5\linewidth}
\begin{center}
\subfigure[\scriptsize{Attention of Task ``Books'' in DA-MTL, Ouptut: Negative}]{
\includegraphics[width=8cm]{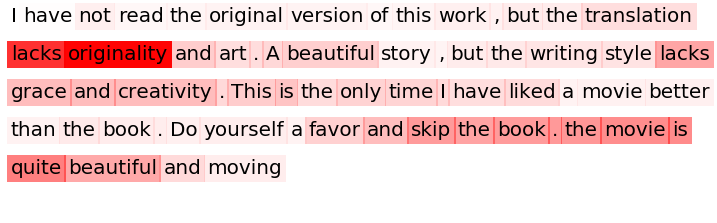}
}
\end{center}%\vspace{-4ex}
\end{minipage}

\end{tabular}
\caption{Attention Distributions  of four classifiers from two models on the same text}
\vspace{-3ex}
\label{fig:attention}
\end{figure}

As shown in Figure \ref{fig:attention}, we also show the attention distributions on a real example selected from the book review dataset. This piece of text involves two domains. The review is negative in the book domain while it is positive from the perspective of movie review. In our SA-MTL model, the ``Books'' review classifier from SA-MTL focus on the negative aspect of the book and evaluate the text as negative. In contrast, the ``DVD'' review classifier focuses on the positive part of the movie and produce the result as positive. In case of DA-MTL, the model first focuses on the two domain words ``book'' and ``movie'' and judge the text is a book review because ``book'' has a higher weight. Then, the model dynamically generates a query $\bq$ and focuses on the part of the book review in this text, thereby finally predicting a negative sentiment.

\subsection{Exp II: Transferability of Shared Sentence Representation }

With attention mechanism, the shared sentence encoder in our proposed models can generate more generic task-invariant representations, which can be considered as off-the-shelf knowledge and then be used for unseen new tasks.

To test the transferability of our learned shared representation, we also design an experiment shown in Table \ref{tab:transfer_result}. The multi-task learning results are derived by training the first $6$ tasks in general multi-task learning. For transfer learning, we choose the last $10$ tasks to train our model with multi-task learning, then the learned shared sentence encoding layer are  kept frozen and transferred to train the first $6$ tasks.

% but not frozen in \textbf{Transfer (not fix)}

\begin{table}[t] \setlength{\tabcolsep}{5pt} \small
\centering
\begin{tabular}{l @{\extracolsep{\fill}} cccc} \hline
                   & SSP-MTL & PSP-MTL & SA-MTL& DA-MTL \\ \hline
Multi-task          & 83.12    & 83.25  & 84.38 & 86.96  \\
Transfer      & 82.54    & 82.58  & 86.50 & 87.67 \\\hline
%Transfer (not fix) & 87.29    & 87.75  & 88.96 & 88.62 \\
\end{tabular}
\caption{Results of first $6$ tasks with multi-task learning and transfer learning}
\vspace{-3ex}
\label{tab:transfer_result}
\end{table}

\vspace{-2ex}

\paragraph{Results and Analysis}

As shown in Table \ref{tab:transfer_result}, we can see that SA-MTL and DA-MTL achieves better transfer learning performances compared to SSP-MTL and PSP-MTL. The reason is that by using attention mechanism, richer information can be captured into the shared representation layer, thereby benefiting the other task.

\subsection{Exp III: Introducing Sequence Labeling as Auxiliary Task}

A good sentence representation should include its linguistic information. Therefore, we incorporate sequence labeling task (such as POS Tagging and Chunking) as an auxiliary task into the multi-task learning framework, which is trained jointly with the primary tasks (the above 16 tasks of sentiment classification). The auxiliary task shares the sentence encoding layer with the primary tasks and connected to a private fully connected layer followed by a softmax non-linear layer to process every hidden state $\bh_t$ and predicts the labels.

%\begin{align}\label{eq:4}
%\hat{y_i} &= \rm{softmax}(W h_t + b),
%\end{align}

\vspace{-2ex}

\paragraph{Dataset}
%As discussed in the previous section, our proposed model can easily incorporate auxiliary sequence labeling tasks such as POS tagging and chunking.
We use CoNLL 2000 \cite{sang2000introduction} sequence labeling dataset for both POS Tagging and Chunking tasks. There are 8774 sentences in training data, 500 sentences in development data and 1512 sentences in test data. The average sentence length is 24 and has a total vocabulary size as 17k.

\vspace{-2ex}

\paragraph{Results}

\begin{table}[t] \setlength{\tabcolsep}{5pt}\small
\centering
\begin{tabular}{l @{\extracolsep{\fill}} rrrr} \hline
              & SSP-MTL & PSP-MTL& SA-MTL & DA-MTL \\ \hline
Origin        & 86.2    & 86.2   & 87.59  &  88.22  \\
+ Chunking    & 86.94   & 86.29  & 88.62  &  88.85  \\
+ POS Tagging & 86.83   & 86.16  & 88.52  &  89.04 \\ \hline
\end{tabular}
\caption{Average precision of multi-task models with auxiliary tasks. %``Origin'' shows the average precision of multi-task models trained with 16 review data sets as introduced in the former section. ``+Chunking'' and ``+POS Tagging'' denotes the average precision on the 16 sentiment classification tasks by using Chunking and POS Tagging as auxiliary task respectively.
}
\vspace{-3ex}

\label{tab:auxiliary_result}
\end{table}

The experiment results are shown in Table \ref{tab:auxiliary_result}. We use the same hyperparameters and training procedure as the former experiments. The result shows that by leveraging auxiliary tasks, the performances of SA-MTL and DA-MTL achieve more improvement than PSP-MTL and SSP-MTL.
%while the performance of PSP-MTL has very limited improvement and even decreases.

\paragraph{Visualization}

\begin{figure}[t]
\begin{tabular}{c}%\hspace{-2ex}
\begin{minipage}[t]{0.5\linewidth}
\begin{center}
\subfigure[\scriptsize{Model trained without Chunking task, Output: Positive}]{
\includegraphics[width=8cm]{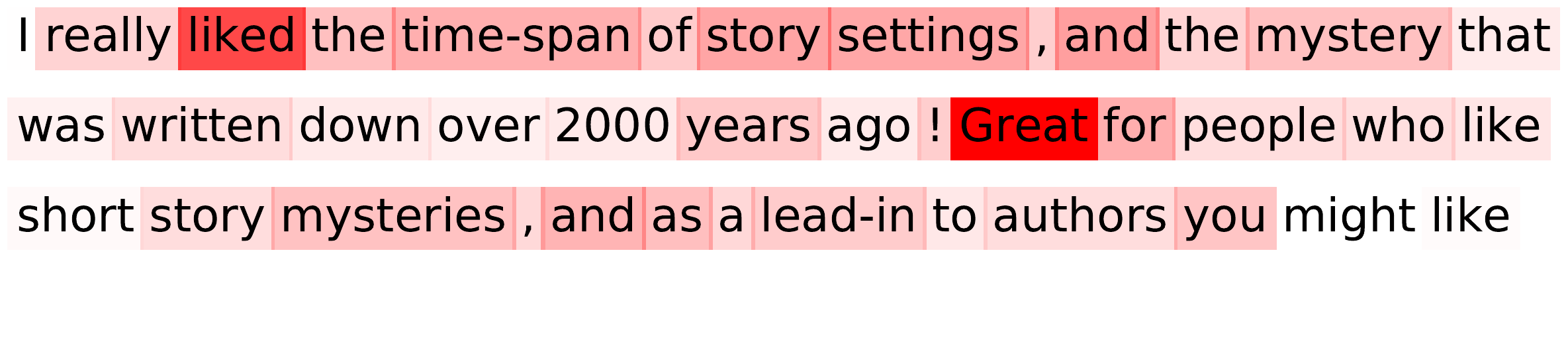}
}
\end{center}%\vspace{-4ex}
\end{minipage}
\\
\begin{minipage}[t]{0.5\linewidth}
\begin{center}
\subfigure[\scriptsize{ Model trained with Chunking task, Output: Positive}]{
\includegraphics[width=8cm]{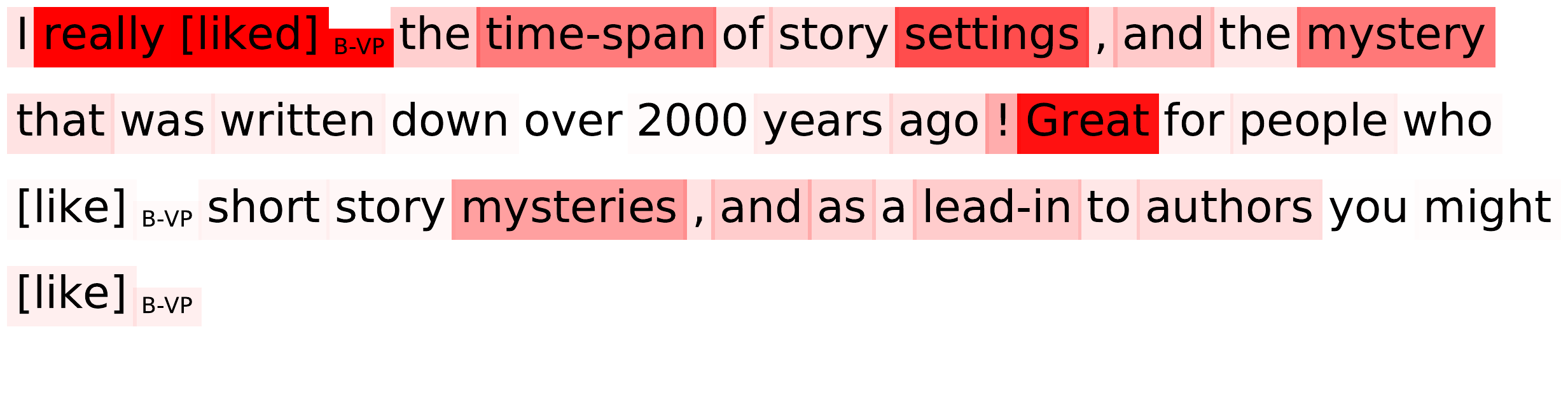}
}
\end{center}%\vspace{-4ex}
\end{minipage}
\\
\begin{minipage}[t]{0.5\linewidth}
\begin{center}
\subfigure[\scriptsize{Model trained without Chunking task, Output: Positive}]{
\includegraphics[width=8cm]{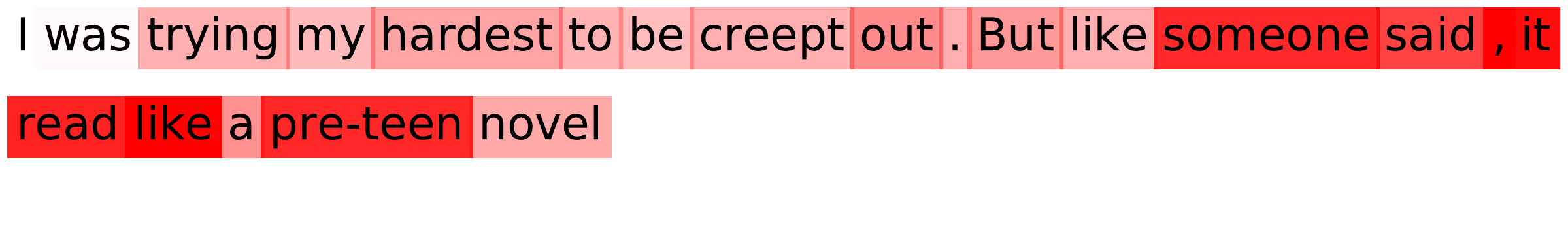}
}
\end{center}%\vspace{-4ex}
\end{minipage}
\\
\begin{minipage}[t]{0.5\linewidth}
\begin{center}
\subfigure[\scriptsize{ Model trained with Chunking task, Output: Negative}]{
\includegraphics[width=8cm]{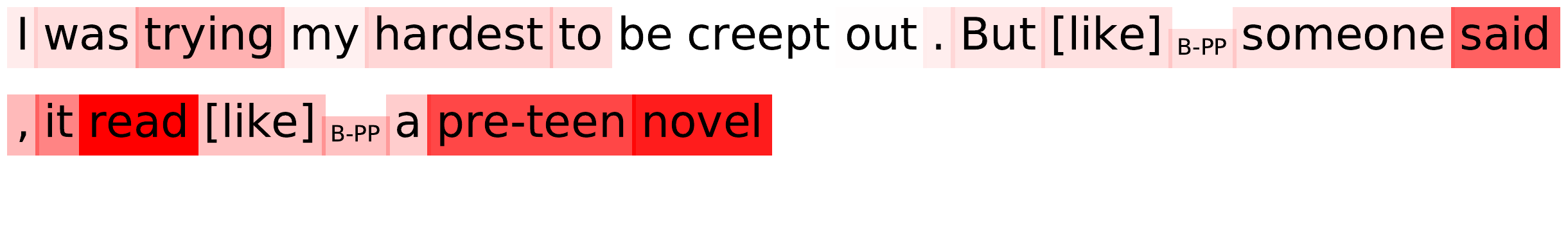}
}
\end{center}%\vspace{-4ex}
\end{minipage}

\end{tabular}
\caption{Attention distributions of two example texts from models trained with and without Chunking task}
\vspace{-3ex}

\label{fig:auxiliary}
\end{figure}

For further analysis, Figure \ref{fig:auxiliary} shows the attention distribution produced by models trained with and without Chunking task on two pieces of texts. In the first piece of text, both of the models attend to the first ``like'' because it represents positive sentiment on the book. The model trained with Chunking task also labels the three ``like'' as 'B-VP' (beginning of verb phrase) correctly. However, in the second piece of text, the same work ``like'' denotes a preposition and has no sentiment meaning. The model trained without Chunking task fails to tell the difference with the former text and focuses on it and produces the result as positive. Meanwhile, the model trained with Chunking task successfully labels the ``like'' as 'B-PP' (beginning of prepositional phrase) and pays little attention to it and produces the right answer as negative. This example shows how the model trained with auxiliary task helps the primary tasks.

\section{Related Work}\label{sec:relate}

%There are two threads of related work. One thread is multi-task learning with neural network.
Neural networks based multi-task learning has been proven effective in many NLP problems \cite{collobert2008unified,glorot2011domain,liu2016recurrent,liu2017adversarial,ruder2017overview} In most of these models, there exists a task-dependent private layer separated from the shared layer.
The private layers play more important role in these models. Different from them, our model encodes all information into a shared representation layer, and uses attention mechanism to select the task-specific information from the shared representation layer.
Thus, our model can learn a better generic sentence representation, which also has a strong transferability.

%There are also a group of work, which uses auxiliary tasks to enhance the representation layer of main task. The auxiliary tasks involve unsupervised tasks, such as sequence autoencoder \cite{dai2015semi} and language model \cite{rei2017semi} and supervised tasks, such as POS Tagging and Chunking \cite{yang2017neural,hashimoto2017joint}. The auxiliary tasks are usually used as a pre-training step.
%\cite{dai2015semi} use them as a pretraining step and find it makes LSTM more stable for long text classification. \cite{rei2017semi} proposed a novel language modeling objective to improve sequence labeling by providing information for surrounding words.
%Instead of using unsupervised auxiliary task, our work is more similar with \cite{yang2017neural} and \cite{hashimoto2017joint} which use supervised sequence labeling task as the auxiliary task.
%Different from \cite{yang2017neural} which uses POS tagging and chunking as a pre-train step,
%Different from these methods, we train the primary and auxiliary tasks jointly and share one single representation layer.

Some recent work have also proposed sentence representation using attention mechanism. \cite{lin2017structured} uses a 2-D matrix, whose each row attending on a different part of the sentence, to represent the embedding.
\cite{vaswani2017attention} introduces multi-head attention to jointly attend to information from different representation subspaces at different positions.
\cite{wang2017learning} introduces human reading time as attention weights to improve sentence representation.
Different from these work, we use attention vector to select the task-specific information from a shared sentence representation. Thus the learned sentence representation is much more generic and easy to transfer information to new tasks.

%\cite{liu2016learning}

%Another thread of work is attention network. Attention network has been widely used in various NLP tasks.
%sentence embedding \cite{lin2017structured}, document classification \cite{yang2016hierarchical} . Different from those works using attention mechanism to do text alignment, we leverage it to extract task-specific information in multi-task learning scheme.

%translation \cite{bahdanau2015neural}, \cite{vaswani2017attention} and question answering \cite{cui2017attn-over-attn}

%\cite{wang2017learning}
%\cite{hahn2016modeling}

\section{Conclusion}\label{sec:conclude}

In this paper, we propose a new information-sharing scheme for multi-task learning, which uses attention mechanism to select the task-specific information from a shared sentence encoding layer.
%The shareable sentence representation is generic and task-invariant to some extent.
We conduct extensive experiments on 16 different sentiment classification tasks, which demonstrates the benefits of our models. Moreover, the shared sentence encoding model can be transferred to other tasks, which can be further boosted by introducing auxiliary tasks.

%In the future, we would like to integrate more tasks, such as language model or machine translation, to learn a better shareable sentence encoding model.

%\bibliographystyle{plainnat}
%\bibliographystyle{aaai}
\bibliographystyle{ijcai18}
\bibliography{attn-mtl,nlp}

\end{document}